\documentclass[lettersize,journal]{IEEEtran}
\usepackage{amsmath,amsfonts}
\usepackage{algorithm}
\usepackage{array}
\usepackage[caption=false,font=normalsize,labelfont=sf,textfont=sf]{subfig}
\usepackage{textcomp}
\usepackage{stfloats}
\usepackage{url}
\usepackage{verbatim}
\usepackage{graphicx}
\usepackage{cite}
\usepackage{booktabs}
\usepackage{xcolor}%
\usepackage{textcomp}%
\usepackage{algorithm}%
\usepackage{algpseudocode}%
\usepackage{amsmath,amssymb,amsfonts}%
\usepackage{amsthm}%
\usepackage{pifont}
\usepackage{bbm}
\usepackage{multirow}%
\begin{document}

\title{SP${ }^3$ : Superpixel-propagated pseudo-label learning for weakly semi-supervised medical image segmentation}


\author{Shiman Li, Jiayue Zhao, Shaolei Liu, Xiaokun Dai, Chenxi Zhang and Zhijian Song
\thanks{Shiman Li is with the School of Basic Medical Science, Fudan University, 200433, China; Shanghai Key Laboratory of Medical Image Computing and Computer Assisted Intervention, 200032, China. (e-mail: smli21@m.fudan.edu.cn).}
\thanks{Jiayue Zhao is with the School of Basic Medical Science, Fudan University, 200433, China; Shanghai Key Laboratory of Medical Image Computing and Computer Assisted Intervention, 200032, China. }
\thanks{Shaolei Liu is with the School of Basic Medical Science, Fudan University, 200433, China; Shanghai Key Laboratory of Medical Image Computing and Computer Assisted Intervention, 200032, China. }
\thanks{Xiaokun Dai is with the Academy for Engineering and Technology, Fudan University, 200433, China.}
\thanks{Chenxi Zhang (Corresponding author) is with the School of Basic Medical Science, Fudan University, 200433, China; Shanghai Key Laboratory of Medical Image Computing and Computer Assisted Intervention, 200032, China. (e-mail: Chenxizhang@fudan.edu.cn).}
\thanks{Zhijian Song (Corresponding author) is with the School of Basic Medical Science, Fudan University, 200433, China; Shanghai Key Laboratory of Medical Image Computing and Computer Assisted Intervention, 200032, China. (e-mail: zjsong@fudan.edu.cn).}
}
\markboth{Journal of \LaTeX\ Class Files,~Vol.~14, No.~8, August~2021}%
{Shell \MakeLowercase{\textit{et al.}}: A Sample Article Using IEEEtran.cls for IEEE Journals}


\maketitle

\begin{abstract}
Deep learning-based medical image segmentation helps assist diagnosis and accelerate the treatment process while the model training usually requires large-scale dense annotation datasets. Weakly semi-supervised medical image segmentation is an essential application because it only requires a small amount of scribbles and a large number of unlabeled data to train the model, which greatly reduces the clinician's effort to fully annotate images. To handle the inadequate supervisory information challenge in weakly semi-supervised segmentation (WSSS), a SuperPixel-Propagated Pseudo-label (SP${}^3$) learning method is proposed, using the structural information contained in superpixel for supplemental information. Specifically, the annotation of scribbles is propagated to superpixels and thus obtains a dense annotation for supervised training. Since the quality of pseudo-labels is limited by the low-quality annotation, the beneficial superpixels selected by dynamic thresholding are used to refine pseudo-labels. Furthermore, aiming to alleviate the negative impact of noise in pseudo-label, superpixel-level uncertainty is incorporated to guide the pseudo-label supervision for stable learning. Our method achieves state-of-the-art  performance on both tumor and organ segmentation datasets under the WSSS setting, using only 3\% of the annotation workload compared to fully supervised methods and attaining approximately 80\% Dice score. Additionally, our method outperforms eight weakly and semi-supervised methods under both weakly supervised and semi-supervised settings. Results of extensive experiments validate the effectiveness and annotation efficiency of our weakly semi-supervised segmentation, which can assist clinicians in achieving automated segmentation for organs or tumors quickly and ultimately benefit patients.

\end{abstract}



\section{Introduction}
Medical Image Segmentation (MIS) is a crucial task in medical image analysis, and holds significant importance for clinical assessment, diagnosis, and treatment \cite{54,wang2023comprehensive}. Recently, with the rapid advancement of deep learning, deep learning-based methods have gradually become mainstream in MIS. U-Net \cite{57} and nn-Unet \cite{58} exhibit impressive learning abilities in MIS, demanding extensive datasets with fine-grained annotations for training.

Nevertheless, acquiring high-quality full annotations for large-scale medical images is often challenging and costly \cite{60,jin2023density,jin2022cold}. In recent years, approaches have been explored to reduce annotation costs, primarily following two strategies: 1) using weakly annotated data (e.g., image-level label, bounding boxes, scribbles, etc.), such as weakly supervised segmentation. Scribble-based learning has gained popularity due to the user-friendly manual annotation, allowing annotators to quickly and accurately label a small pixel set \cite{11,chen2023adversarial,zhang2024attention}. 2) utilizing unlabeled data, such as semi-supervised segmentation. Semi-supervised segmentation enhances segmentation performance by mining information in unlabeled data to address limited annotation challenges \cite{63,xu2022gct,wang2022mmatch}. 

\begin{figure}[t]
    \centering
    \includegraphics[width=1.0\columnwidth]{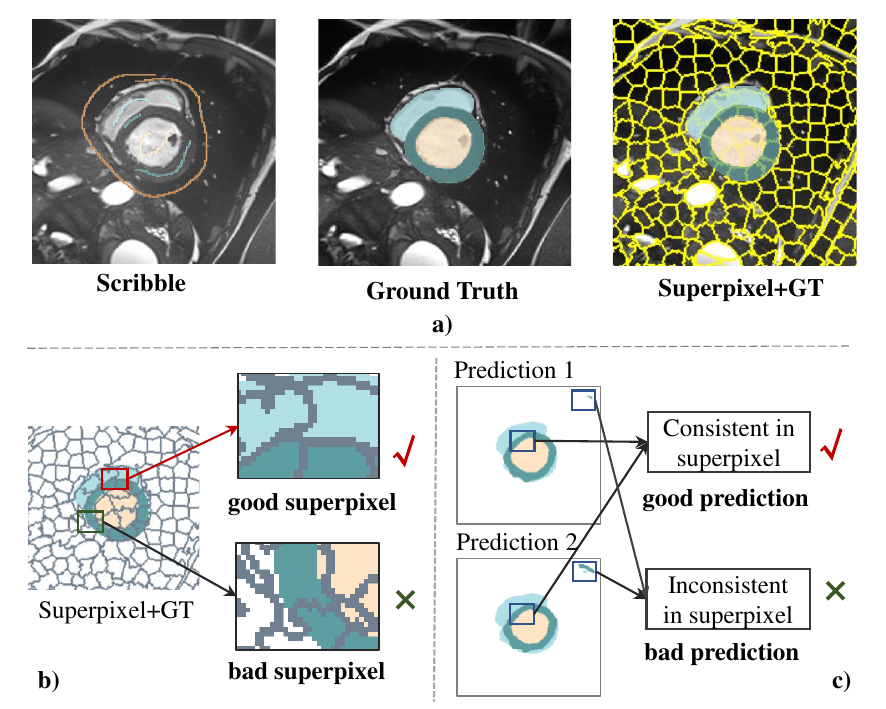} 
    \caption{a) Sample images of scribbles, ground truth, and superpixels. b) Illustration of superpixel quality. c) Illustration of prediction quality.}
    \label{fig1}
    \end{figure}

To cut annotation costs further, there's a focus on merging different learning methods for mixed supervision. For instance, Weakly Semi-Supervised Segmentation (WSSS) trains networks with limited weakly annotated data and abundant unlabeled data to achieve notable performance \cite{30,31}. Gao et al. \cite{65} utilized scribbles as the weak annotation and employs multi-task learning for distinct supervision of different data. However, due to the large quantity of unlabeled data compared to weak annotation data, the inconsistent supervision results in over-guidance from unlabeled data, causing suboptimal representations during the training process \cite{26}. Therefore, we employ pseudo-label learning to impose unified supervision on both types of data, mitigating inconsistent supervision issues.

However, in WSSS, establishing an efficient pseudo-label learning method faces two main challenges. \textbf{Firstly, insufficient supervisory information} leads to ambiguous boundaries and the defective appearance of pseudo-labels. \textbf{Secondly, noisy pseudo-labels} brought by weakly labeled data and vast unlabeled data may accumulate errors and lead to unstable segmentation. To tackle these issues, we propose a superpixel-propagated pseudo-label learning approach to promote weakly semi-supervised medical image segmentation.

As illustrated in Figure \ref{fig1}(a), sparse annotations like scribbles lack complete target regions and precise boundaries compared to dense annotations. Conversely, superpixels and ground truth exhibit regional and boundary-wise consistency. Pixels within a superpixel share textual similarity usually corresponding to the same semantic category and superpixels' boundaries often exhibit pronounced grayscale gradients similar to ground truth. \textit{Therefore, we aim to utilize the image's prior information (texture and boundary) contained within superpixels to boost the pseudo-label supervision, complementing insufficient supervision.} However, the superpixels may contain some noise that is inconsistent with ground truth, so it’s important to assess such bad superpixels during training. As shown in Figure \ref{fig1}(b), the superpixel quality can be distinguished by the class proportion, for poor superpixel with lower class proportion may introduce boundary noise. Given this, we would like to filter high-quality superpixels for pseudo-label refinement, propagating beneficial boundary information in superpixels.

Moreover, the high noise in pseudo-labels produced by WSSS models remains non-negligible even after high-quality superpixel refinement. A way to discriminate the potential noise in prediction is based on the consistency between multiple outputs, as shown in Figure \ref{fig1}(c). A consistent region often corresponds to a simple region with a lower probability of bad prediction, and bias tends to occur in difficult regions with inconsistent predictions. \textit{To further mitigate the effect of noisy pseudo-labels, we propose superpixel-level uncertainty based on the consistency of predictions to guide and stabilize pseudo-label supervision.} 

In this paper, to address the above issues, a comprehensive superpixel-propagated pseudo-label learning framework are proposed. Our superpixel-propagated pseudo-label learning method consists of three components: superpixel-based scribble expansion; dynamic threshold filtering superpixels for pseudo-label refinement; and superpixel-level uncertainty guidance. Specifically, to fully utilize limited label information of scribble, we employ a scribble expansion approach that expands scribble labels to similar adjacent pixels bounded by superpixels to obtain superpixel-level dense annotations. Subsequently, to improve pseudo-label quality, we use high-resolution superpixels to refine pseudo-labels by compensating image details like boundaries. The dynamic threshold adaptively selects simple and consistent superpixels for relabeling during the training process. Additionally, we utilize the difference between predictions from two decoders within a superpixel to gauge superpixel-level uncertainty. Then, introducing superpixel-level uncertainty as weights for pseudo-label supervision directs the network's attention towards more dependable pseudo-label regions, thus augmenting model stability. To recap, the main contributions are as follows:

\begin{itemize}
    \item To alleviate the insufficient and inconsistent supervision in WSSS, a superpixel-propagated pseudo-label learning is proposed to use the superpixel for supplementary information and union pesudo label learning for both weakly annotated and unlabeled data.
    \item A dynamic threshold filtering strategy is employed to ensure the propagation of beneficial superpixel information during training.
    \item The superpixel-level uncertainty is applied to guide pseudo-label learning and alleviates the bias caused by imprecise pseudo-label predictions.
    \item Our WSSS method achieves superior performance in two public datasets and can be applied to other annotation-efficient learning.

\end{itemize}

\section{Related Works}
\subsection{Weakly Supervised Learning in MIS}

Weakly supervised learning simplifies the manual annotation task to image-level labels \cite{2}, bounding boxes \cite{4}, points \cite{6}, and scribbles \cite{7,9}. In this paper, we focus on scribble-based learning. Some approaches generated initial dense annotations by relabeling adjacent pixels through methods like region growing \cite{11} and graph-based techniques \cite{10}. These steps, however, are time-intensive and introduce label noise. So, conditional random fields (CRF) were integrated either as post-processing \cite{11} or trainable layers \cite{10}, enhancing segmentation results but necessitating multi-stage training and updates.

Lee et al. \cite{17} proposed a self-training-based pseudo-label learning and selected reliable pseudo-labels by confidence thresholding for network training. Luo et al. \cite{18} dynamically mixed two predictions from dual branches to generate pseudo-labels for supervision. Zhang et al. \cite{74} incorporated a mixup strategy and cycle consistency for effective scribble-based learning.  Liu et al. \cite{19} presented an uncertainty-aware self-ensemble approach to produce reliable predictions for supervision and impose regularization through transformation consistency. Additionally, other studies \cite{21,22} assessed segmentation quality and encouraged the generated segmentation to be closer to the ground truth segmentation through adversarial learning, but required additional dense annotations.

\begin{figure*}[t]
\centering
\includegraphics[width=0.83\paperwidth]{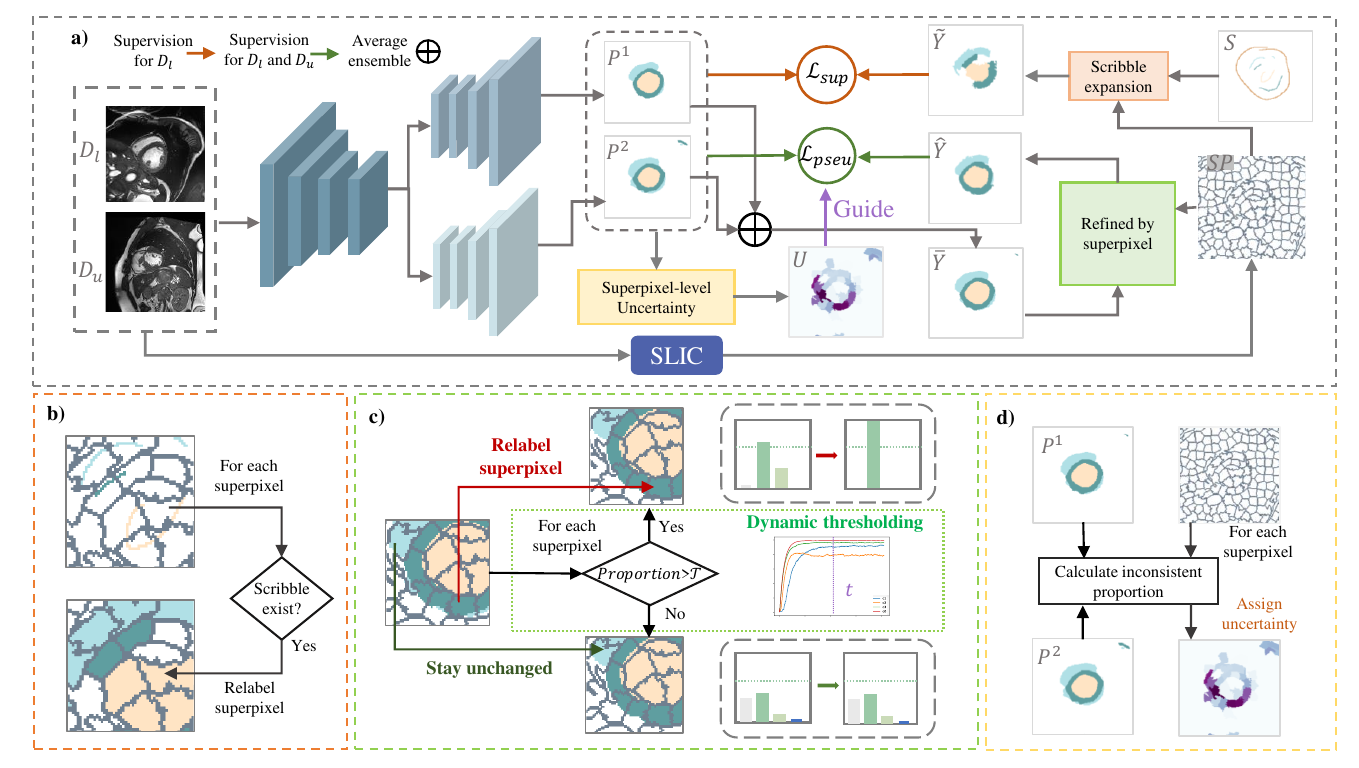} 
\caption{a) An overview of the proposed method. b) Illustration of superpixel-based scribble expansion. c) Illustration of pseudo-label refinement with superpixel filtered by dynamic threshold. d) Illustration of superpixel uncertainty assessment.}
\label{fig2}
\end{figure*}

\subsection{Semi-Supervised Learning in MIS}
The dominant semi-supervised learning for MIS can be categorized into pseudo label-based and consistent regularization-based methods. Pseudo label-based methods assign pseudo-labels to unlabeled data, which acts as supervision alongside labeled data \cite{lu2024self,}. The quality of these pseudo-labels is optimized iteratively, with ensemble techniques often boosting their accuracy \cite{37}. Furthermore, due to the existence of the pseudo-labeling noise, it is crucial to assess and differentiate pseudo-label quality. Certain approaches utilized confidence \cite{34} or prediction consistency \cite{36} as reliability indicators, masking out unreliable pixels in the pseudo-label loss. Zhou et al. \cite{35} even employed dropout strategies to estimate uncertainty, guiding pseudo-label learning and improving noise resistance. The consistency regularization approach enforces consistent results under small perturbations \cite{43,cao2022adversarial}, as regularization terms for unlabeled data exploration. Given the potential outliers induced by diverse perturbations \cite{41}, incorporating uncertainty considerations when enforcing consistency becomes valuable. Yu et al. \cite{47} estimated uncertainty maps via Monte Carlo dropout \cite{75}, though computational load increased significantly. Wu et al. \cite{73} measured uncertainty using multiple outputs from slightly different decoders. Luo et al. \cite{45} directly calculated KL divergence between multi-scale predictions as uncertainty, minimizing it for enhancing model robustness. However, multiple decoders or multi-scale pyramid models entail greater resource consumption.

\subsection{Mixed supervised learning in MIS}
Mixed supervised learning aims to strike a balance between network performance and annotation cost by combining various annotation types. A common way is to supplement limited dense annotated data with abundant weakly annotated data, such as image class labels \cite{29}, bounding boxes \cite{30}, and scribbles \cite{25,26}. However, simply treating two types of data equally only contributes to limited improvement, due to the underutilization of weak annotations \cite{27}. Existing methods employed different treatments through multi-stage training \cite{32}, multi-branch networks \cite{25,26}, and multi-task learning \cite{29,30} to further explore weakly annotated data. Yet, the performance of these methods relies on full annotated data, and the learning from weak annotations is usually guided by full annotations. Moreover, some studies explored WSSS using minimal weak annotations alongside abundant unlabeled data, and limited annotations posing greater challenges. Gao et al. \cite{65} proposed a SOUSA method based on a teacher-student framework to unify weakly semi-supervised segmentation, leveraging both scribble-annotated and unlabeled data via multi-task learning. They applied geodesic distance map regression loss for sparse annotations supervision and multi-view mapping consistency loss for unlabeled data supervision. Nevertheless, multi-task learning may over-rely on the rich unlabeled data, neglecting guidance from weak annotation information. This supervision inconsistency is due to the sample imbalance and will limit model performance \cite{27}. In this paper, we perform superpixel-propagated pseudo-label learning for MIS, training scribble-annotated data and unlabeled data with unified pseudo-label supervision to address supervision inconsistency. And we leverage superpixel propagation to alleviate insufficient supervision under high noise, thus optimizing model performance.

\subsection{Superpixels in MIS}
Superpixel segmentation is a way of over-segmenting an image by composing pixels that are adjacent and similar to small regions as superpixels \cite{felzenszwalb2004efficient,sasmal2023aquila,dhal2023chaotic}. Superpixels are traditionally generated by graph-based \cite{ren2003learning} and clustering-based \cite{ray2023superpixel,sasmal2023survey} unsupervised algorithms. The generated superpixels preserve low-level image representations that are useful for image segmentation and help to extract boundary information of image objects. Hence, some current works incorporate superpixels to further improve segmentation performance in deep learning. Ouyang et al. \cite{ouyang2020self} use superpixels for self-supervised learning to help improve the performance of network few-shot segmentation. Wang et al. \cite{wang2022separated} introduce superpixels for dense contrast learning to improve multi-organ representation. Thompson et al. \cite{thompson2022pseudo} merge the pseudo labels of all similar superpixels in semi-supervised learning. Li et al. \cite{li2021superpixel} use the internal consistency of superpixels to design regularisation terms to achieve learning from noisy labels. However, none of the above methods adequately consider the mis-segmentation in superpixels and no methods have yet been developed to evaluate the quality of different superpixels and handle bad superpixels to avoid the effects of their noise. In contrast, our method filters high-quality superpixels through dynamic thresholding and assigns different weights to different superpixel regions through superpixel-level uncertainty, which helps to ensure the utilization of beneficial superpixels and avoid the propagation of erroneous superpixel information.

\section{Method}

\subsection{Problem setup and overview of framework}
In the weakly semi-supervised segmentation task, a small number of weakly labeled data $D_l$ and a large number of unlabeled data $D_u$ are used for model training, and the obtained model $f\left(\theta\right)$ can segment the test data.
In this paper, the scribbles are used as the weak annotations, and the labeled dataset $D_l=\left\{\left(X^b, S^b\right)\right\}_{b=1}^{N^l}$ contains $N^l$ samples. $X$ denotes the medical image and $S=\left\{s_k,c_k\right\}$ denotes its corresponding scribble annotation, where $s_k$ denotes the pixel set of scribble $k$, and $0 \leq c_k \leq C$ is the class of the scribble. The unlabeled dataset $D_u=\left\{X^b\right\}_{b=1}^{N^u}$ has $N^u$ images without annotations.

An encoder-decoder network architecture is employed as the backbone, as shown in Figure \ref{fig2}. The framework comprises an encoder and two slightly different decoders (one with dropout and one without). The framework works in conjunction with three operations: scribble expansion, pseudo-label refinement, and superpixel-level uncertainty guidance, whose details are shown in the subfigure in Figure \ref{fig2} respectively. Here, the algorithm of our method is exhibited in Algorithm \ref{alg1}. First of all, the superpixels of all images are generated offline by Simple Linear Iterative Clustering (SLIC). During training, the network samples both weakly labeled data $D_l$ and unlabeled data $D_u$ as input simultaneously. For each input image X, the network outputs two predictions $P^1$ and $P^2$, and applies the average ensemble strategy to obtain pseudo label $\bar{Y}$. For the labeled data, the scribble annotations are expanded by superpixels to generate expanded scribbles $\tilde{Y}$, which is used to calculate the partial supervised loss $\mathcal{L}_{sup}$. Additionally, for both weakly labeled and unlabeled data, their pseudo labels $\bar{Y}$ are refined by superpixels selected by the dynamic threshold to obtain refined pseudo labels $\hat{Y}$. Besides, two predictions from the network can be used to calculate superpixel-level uncertainty, which is used to guide the pseudo-label supervision loss $\mathcal{L}_{pseu}$ by reweighting the superpixel region. Finally, the model parameters are optimized by the total loss of supervised loss and pseudo-label loss in each iteration. More methodological details can be found in the subsequent subsections. 

\begin{algorithm}[]
    \caption{The overall algorithm of the proposed method.}
    \label{alg1}
    \textbf{Input}: $D_l, D_u$ : labeled and unlabeled dataset.\\
    \textbf{Output}: model's parameter $\theta$.
    \begin{algorithmic}[1] 
    \State  Obtain superpixels $S P$ for each image $X \in D_l \cup D_u$:

$SP \leftarrow SLIC(X)$
    \While{stopping criterion not met}
    \State Sample $\left\{\left(X_b, S_b\right)\right\}_{b=1}^{\mu B} \sim D_l$ and $\left\{X_b\right\}_{b=1}^{(1-\mu) B} \sim D_u$

    \State Obtain two predictions $P^1, P^2$ and average them as pseudo label $\bar{Y}$:

	$\bar{Y}=\operatorname{argmax}\left(\left(P^1+P^2\right) * 0.5\right)$

    \State Obtain expended scribble $\tilde{Y}$ for labeled image with scribbles $S$:

	$\tilde{Y}=\varphi_1(S P, S)$
    
    \State Compute supervised loss $\mathcal{L}_{S U P}$ of the labeled images:

	$\mathcal{L}_{\text {SUP }}=\frac{1}{\mu B} \sum_b^{\mu B} \mathcal{L}_{\text {sup }}\left(\tilde{Y}^b, P_b^1, P_b^2\right)$
    \State Refine pseudo labels by the superpixels filtered by dynamic threshold:

	$\hat{Y}=\varphi_2(\bar{Y}, S P)$
    \State Obtain superpixel-level uncertainty map $U$:

	$U=\varphi_3\left(P^1, P^2, S P\right)$
    \State Compute pseudo label loss guided by superpixel-level uncertainty $\mathcal{L}_{P S E U}$:

	$\mathcal{L}_{\text {PSEU }}=\frac{1}{B} \sum_b^B \mathcal{L}_{\text {pseu }}\left(P_b^1, P_b^2, \hat{Y}^b, U^b\right)$
    \State Compute the total loss and the gradient of loss to update the network parameters $\theta$ by the backpropagation.

	$\mathcal{L}_{\text {TOTAL }}=\mathcal{L}_{S U P}+\mathcal{L}_{\text {PSEU }}$
    \EndWhile
    \State \textbf{return} $\theta$
    \end{algorithmic}\label{alg1}
\end{algorithm}

\subsection{Superpixel-based Scribble Expansion}
We employ the conventional SLIC algorithm for superpixel generation. For a single image $X$, SLIC partitions image into $n$ superpixels $SP=\left\{sp_j \right\}_{j=1}^n$. Each superpixel $sp_j=\left\{x_i \right\}_{i=1}^{M_j }$ contains $M_j$  similar adjacent pixels $x_i$. Here, we set $n$ to a relatively large value to ensure each superpixel is smaller than the size of the target and exclude the case of superpixel with multi-class scribbles.
Since the scribbles only annotate limited pixels, we leverage pixel similarity within superpixel to expand the scribbles. We assign superpixel with the class of scribble it contains, and obtain the expanded scribble $\tilde{Y}=\varphi_1\left(\tilde{y}_{i j} \mid S, SP\right)$. The corresponding mapping relationship is formulated as:

\begin{equation}
\varphi_1\left(\tilde{y}_{i j}\right)=\left\{\begin{array}{c}
c_k, \text { if } s p_j \cap s_k \neq \emptyset \\
\text { none, else }
\end{array}\right.
\end{equation}

Specifically, if scribble $s_k$ exists within superpixel $sp_j$, the label of pixels $\tilde{y}_{ij}$ which are contained in the $sp_j$ region is assigned the class $c_k$. Otherwise, no assignment is performed. Thus, we can leverage the expanded scribbles to provide partial supervision for two predictions $P^1$ and $P^2$ from dual branches:
\begin{equation}
\mathcal{L}_{\text {sup }}=\left(\mathcal{H}_p\left(\tilde{Y}^b, P_b^1\right)+\mathcal{H}_p\left(\tilde{Y}^b, P_b^2\right)\right) * 0.5 
\end{equation}
\begin{equation}
\mathcal{H}_p=\sum_c^C \sum_{i \in \Omega} y_i^c \log p_i^c
\end{equation}
where, $B$ denotes the batch size, $\mu$ is the proportion of weakly annotated data in the batch. $\mathcal{H}_p$ is the partially cross-entropy loss, where the cross-entropy of prediction $p_i^c$ and label $y_i^c$ only calculate for the pixels in region $\varOmega$. When computing $\mathcal{L}_{\text {sup }}$, $\varOmega$ represents the set of superpixels containing scribbles.

\subsection{Pseudo-label refinement with superpixel filtered by dynamic threshold}
To get full supervision for the whole image, we apply pseudo-label learning with superpixel-based refinement. The pseudo label $\bar{Y}$ is generated by averaging two predictions of network:  
\begin{equation}
\bar{Y}=\operatorname{argmax}\left(\left(P^1+P^2\right) * 0.5\right)
\end{equation}

To address blurred boundaries of pseudo-labels, we leverage superpixels to help capture edges for pseudo-label refinement. We filter superpixels by the proportion of the dominant class to relabel high-quality superpixels, thereby obtaining the refined pseudo-labels $\hat{Y}=\varphi_2\left(\hat{y}_{i j} \mid \bar{Y}, SP\right)$. We relabel superpixels with their maximum class when the proportion exceeds the threshold, and otherwise keep the original prediction. This process is visually explained in Figure 4. The mapping function $\varphi_2$ is defined as:
\begin{equation}
\varphi_2\left(\hat{y}_{i j}\right)=\left\{\begin{array}{c}
\operatorname{argmax}\left(\bar{y}_j\right), \text { if } \psi\left(\bar{y}_{ij}\right)>\mathcal{T} \\
\bar{y}_{ij}, \text { else }
\end{array}\right. \\
\end{equation}
\begin{equation}
\psi\left(\bar{y}_j\right)=\sum_{i \in sp_j} \mathbbm{1}\left(\bar{y}_{i j}==\operatorname{argmax}\left(\bar{y}_j\right)\right) / M_j
\end{equation}
where $\psi(\bar{y}_j )$ denotes the proportion of pixels with dominant class in superpixel $sp_j$, and $\mathcal{T}$ is the threshold.
Considering the influence of the threshold on the learning status, we design a dynamic threshold for filtering inspired by the adaptive thresholding in \cite{49}. To better exploit superpixel information, the dynamic threshold uses an exponentially moving average (EMA) strategy performing an increasing trend during training, which ensures reliable superpixels are relabeled and unreliable ones are gradually ignored as training progresses. Furthermore, given the class imbalance, we define class-specific thresholds as follows:
\begin{equation}
\mathcal{T}_t(c)=\left\{\begin{array}{c}
\tau_0, \text { if } t=0 \\
\lambda \tau_{t-1}(c)+(1-\lambda) \frac{1}{B} \sum_{b=1}^B \max (\psi(c)), \text { else }
\end{array}\right.
\end{equation}
where $\mathcal{T}_t(c)$ denotes the threshold for class $c$ at the $t$-th time step (iteration), $\tau_0$  indicates the initial threshold and $\lambda$ is the momentum decay of EMA to dynamically update the thresholds. $\psi(c)$ is the class proportion of class $c$.

\subsection{Superpixel-level Uncertainty Guidance}
In order to address the influence of noisy pseudo-label, we utilize superpixel-level uncertainty to guide pseudo-label supervision. Considering that the uncertainty is associated with consistency of two predictions, we compute the ratio of the number of pixels with different predictions within superpixel to assess the superpixel-level uncertainty of the pseudo label:
\begin{equation}
\varphi_3\left(u_{j}\right)=\frac{\sum_{i \in sp_j} \mathbbm{1}\left(\operatorname{argmax}\left(p^1_i\right) \neq \operatorname{argmax}\left(p^2_i\right)\right)}{M_j}
\end{equation}

\begin{equation}
\mathcal{L}_{\text {pseu }}=\mathcal{L}_{\text {wdice }}\left(P^1, \hat{Y} \mid U\right)+\mathcal{L}_{\text {wdice }}\left(P^2, \hat{Y} \mid U\right) 
\end{equation}
\begin{equation}
\mathcal{L}_{\text {wdice }}(P, Y)=1-\frac{2 \sum W * P * Y}{\sum P * W+\sum Y * W} 
\end{equation}
\begin{equation}
w_{i j}=e^{-u_j}
\end{equation}
where $\mathcal{L}_{wdice}$ indicates weighted dice loss, $W$ is the weight map based on superpixel-level uncertainty, composed of a union of pixel weight $w_{i j}$. Guided by the superpixel-level uncertainty, the network can focus more on learning from higher quality pseudo-labels. 
Finally, the complete training loss function is defined as:
\begin{equation}
\mathcal{L}_{\text {total }}=\mathcal{L}_{\text {sup }}+\mathcal{L}_{\text {pseu }}
\end{equation}

\section{Experiment}
\subsection{Data and implementation details}
\textbf{Dataset:} We evaluated our method using two publicly available datasets. The ACDC (Automated Cardiac Diagnosis Challenge) \cite{67} dataset consists of cine-MR images from 100 patients. Each image is manually annotated with scribbles for left ventricle (LV), right ventricle (RV), and myocardium (MYO) structures, following the annotations provided in previous work \cite{66}. In ACDC experiments, we performed 5-fold cross-validation with 80 samples for training and 20 samples for testing in each fold. The BraTS2019 dataset \cite{69} consists of 335 multimodal MR images of brain tumors, and we used the preprocessed FLAIR images along with corresponding whole tumor labels from \cite{45} for our segmentation task. To generate corresponding scribble annotations, we used the method introduced in previous work \cite{70} to simulate scribble annotations. In BraTS2019 experiments, we utilized 250, 25, and 60 samples as the training, validation, and test sets, respectively. In WSSS and semi-supervised segmentation, the labeled samples are randomly selected from the training set.

\textbf{Training Details:} Our method was implemented using the PyTorch framework \cite{71} and all experiments were executed on an NVIDIA 3090 GPU. We evaluated our approach on both 2D and 3D image segmentation. For ACDC and BraTS2019 datasets, we adopted U-Net and 3D U-Net as the backbone for encoding and decoding, respectively. During network training, we used stochastic gradient descent (SGD) optimizer with a momentum of 0.9 and weight decay of 0.0001. Learning rates are set to 0.01 for 2D segmentation and 0.1 for 3D segmentation, respectively.

\textbf{Evaluation:} Extensive experiments were conducted to evaluate the performance of our SP${ }^3$ method using four metrics of Dice coefficient, Jaccard Index (JI), 95\% Hausdorff Distance (95HD), and Average Surface Distance (ASD).

\subsection{Comparison with the state-of-the-art}
\subsubsection{Main result}

To validate the effectiveness of our proposed SP${ }^3$ method, we conducted a comparative evaluation against a series of state-of-the-art methods. 
All experiments were performed with the same settings to ensure a fair comparison. Specifically, 
we constructed two baseline methods that do not utilize unlabeled data as references: Sparse Anno. and Dense Anno.
indicate training with labeled data with scribble and full annotations respectively. We compared our method with the leading WSSS methods SOUSA. Due to the lack of WSSS methods for comparison, we adapted advanced weakly supervised methods (S2L, USTM) and semi-supervised (ICT, URPC) methods to the WSSS setting for comparison. For the pseudo-label-based S2L method, we assign pseudo-labels to unlabeled data and impose pseudo-label supervision, the same as scribble-labeled data. As for USTM, we extend its consistency concept to the utilization of unlabeled data, calculating transformation consistency loss for the predictions of unlabeled data from both branches. In the case of semi-supervised methods, we replace the fully dense annotation with weakly annotation to compute partial cross-entropy loss. Simultaneously, we also apply their consistency supervision for unlabeled data to the supervision of scribble-annotated data, adapting the ICT and URPC to the WSSS setting.

\begin{table*}[!t]
    \scriptsize 
    \centering
    \caption{Comparison results of different methods under weakly semi-supervision on ACDC dataset.}
    \begin{tabular}{c|cccc|cccc}
    \toprule
    Method              & Dice          & JI            & 95HD        & ASD         & Dice          & JI            & 95HD         & ASD        \\ \midrule
    labeled   ratio     & \multicolumn{4}{c|}{10\%}                                 & \multicolumn{4}{c}{20\%}                                  \\ \midrule
    Sparse   Anno. (LB) & 0.5232±0.0781 & 0.3754±0.0730 & 116.54±9.03 & 53.55±9.01  & 0.5774±0.0238 & 0.4349±0.0215 & 107.78±12.49 & 43.93±6.88 \\
    Dense   Anno. (UB)  & 0.6824±0.0584 & 0.5709±0.0632 & 16.93±3.39  & 3.74±0.96   & 0.7579±0.0540 & 0.6511±0.0636 & 12.10±3.63   & 2.53±0.96  \\
    S2L\_WSSS \cite{17}           & 0.6383±0.0455 & 0.5043±0.0378 & 43.10±20.37 & 14.96±6.46  & 0.6657±0.0700 & 0.5370±0.0766 & 33.24±17.94  & 10.16±5.48 \\
    USTM\_WSSS \cite{19}          & 0.6814±0.0476 & 0.5526±0.0519 & 64.41±12.65  & 19.49±5.26 & 0.7135±0.0487 & 0.5810±0.0514 & 57.68±4.24 & 16.72±2.30 \\
    URPC\_WSSS \cite{18}         & 0.6805±0.1208 & 0.5498±0.1218 & 48.74±30.06 & 18.30±18.42 & 0.7267±0.0766 & 0.6228±0.0787 & 13.09±3.76  & 2.31±0.71 \\
    ICT\_WSSS \cite{ICT}          & 0.7010±0.0764 & 0.5790±0.0781 & 18.83±4.59  & 6.35±2.38   & 0.7344±0.073  & 0.6153±0.0781 & 33.50±19.00  & 11.91±9.02 \\
    SOUSA \cite{65}             & 0.7224±0.0602 & 0.5980±0.0666 & 37.85±20.11 & 12.98±8.78  & 0.7703±0.0808 & 0.6519±0.0899 & 39.10±16.41  & 12.78±8.72 \\
    \textbf{Ours}            & \textbf{0.7937±0.0306} & \textbf{0.6740±0.0377} & \textbf{13.11±1.56}  & \textbf{3.89±0.52}   & \textbf{0.8186±0.0439} & \textbf{0.7127±0.0531} & \textbf{7.92±2.79}   & \textbf{2.26±0.75}  \\ \bottomrule
    \end{tabular}\label{table1}
\end{table*}

\begin{figure*}[t]
    \centering
    \includegraphics[width=0.9\textwidth]{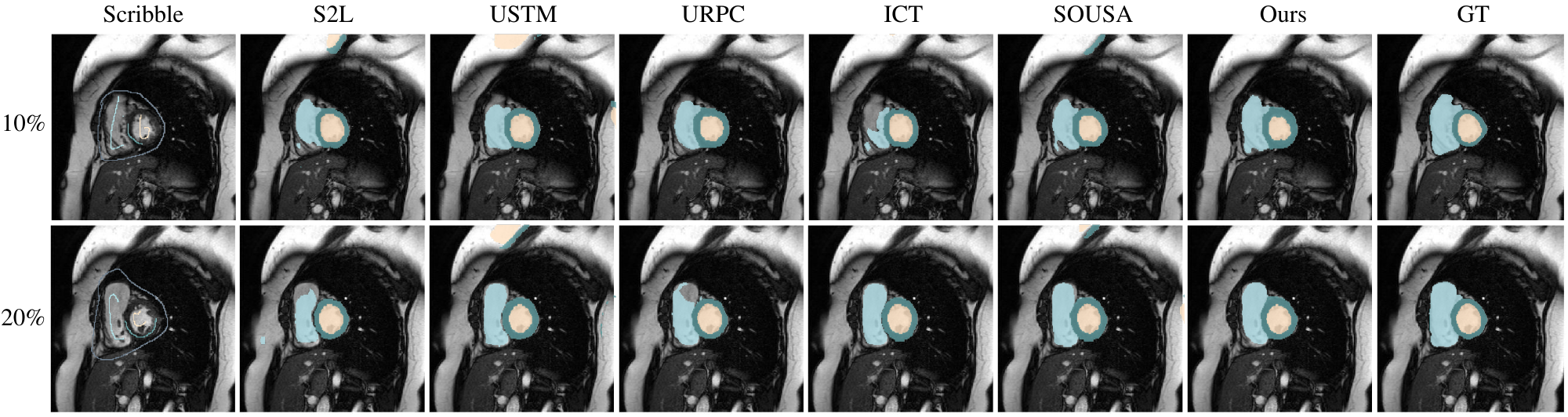} 
    \caption{Visual comparison of weakly semi-supervised segmentation results for the ACDC datasets under different labeled ratios.}
    \label{fig3}
    \end{figure*}
\begin{table*}[!t]
    \scriptsize 
    \centering
    \caption{Comparison results of different methods under weakly semi-supervision on BraTS2019 dataset.}
    \begin{tabular}{c|cccc|cccc}
    \toprule
    Method              & Dice          & JI            & 95HD        & ASD         & Dice          & JI            & 95HD        & ASD         \\ \midrule
    labeled   ratio     & \multicolumn{4}{c|}{5\%}    & \multicolumn{4}{c}{10\%}     \\ \midrule
    Sparse Anno. (LB) & 0.5870±0.2477 & 0.4592±0.2515 & 77.46±35.55 & 35.40±23.99 & 0.6042±0.2536 & 0.4798±0.2603 & 72.62±42.23 & 32.26±24.50 \\
    Dense Anno. (UB)  & 0.7848±0.1854 & 0.6777±0.2109 & 18.59±20.89 & 4.71±5.95   & 0.8093±0.1626 & 0.7057±0.1926 & 15.47±20.75 & 2.50±2.60   \\
    S2L\_WSSS \cite{17}          & 0.7199±0.1745 & 0.5898±0.2029 & 41.96±32.13 & 15.43±15.57 & 0.7284±0.1791 & 0.6017±0.2078 & 42.87±35.55 & 16.46±16.91 \\
    USTM\_WSSS \cite{19}         & 0.6648±0.2185 & 0.5366±0.2379 & 49.72±10.99 & 21.34±20.13 & 0.7288±0.1738 & 0.6011±0.2048 & 40.96±30.87 & 15.76±15.05 \\
    URPC\_WSSS \cite{18}         & 0.6910±0.1968 & 0.5604±0.2182 & 39.41±28.29 & 14.73±14.10 & 0.7256±0.1870 & 0.6002±0.2121 & 34.95±28.96 & 11.83±13.07 \\
    ICT\_WSSS \cite{ICT}          & 0.7012±0.1823 & 0.5686±0.2059 & 53.19±31.32 & 19.40±15.31 & 0.7407±0.1677 & 0.6147±0.2006 & 36.66±34.66 & 13.84±16.21 \\
    SOUSA \cite{65}               & 0.7320±0.1906 & 0.6100±0.2192 & 36.36±32.27 & 14.65±16.98 & 0.7525±0.1629 & 0.6290±0.1982 & 33.40±31.78 & 12.34±15.54 \\
    \textbf{Ours}                & \textbf{0.7506±0.1810} & \textbf{0.6300±0.2042} & \textbf{32.57±32.63} & \textbf{10.09±11.02} & \textbf{0.8064±0.1592} & \textbf{0.7014±0.1946} & \textbf{15.24±20.85} & \textbf{4.72±9.22}   \\ \bottomrule
    \end{tabular}\label{table2}
\end{table*}

\textbf{ACDC dataset}
From Table \ref{table1}, all methods demonstrate improvements over the lower bound using only sparse annotations, confirming that learning from unlabeled data can yield model gains. SOUSA method outperforms other WSSS methods adopted from weakly supervised methods (S2L, USTM) and semi-supervised methods (URPC, ICT) and achieves suboptimal results. Our method achieves the best performance in Dice, JI, 95HD, and ASD compared to other methods, with 0.7937 and 0.8186 Dice scores on 10\% and 20\% labeled ratios respectively. Our method fully leverages the scribble annotations and unlabeled data, achieving superior Dice scores over the upper bound utilizing dense annotations by 0.1113 and 0.0607 on two label ratios respectively. Higher improvement above the upper bound and other comparative methods with lower label ratios show our method can resist noise better. The visual comparison results are shown in Figures \ref{fig3}, in which our method gets the optimal segmentation with more accurate boundaries.

\begin{figure*}[t]
\centering
\includegraphics[width=.92\textwidth]{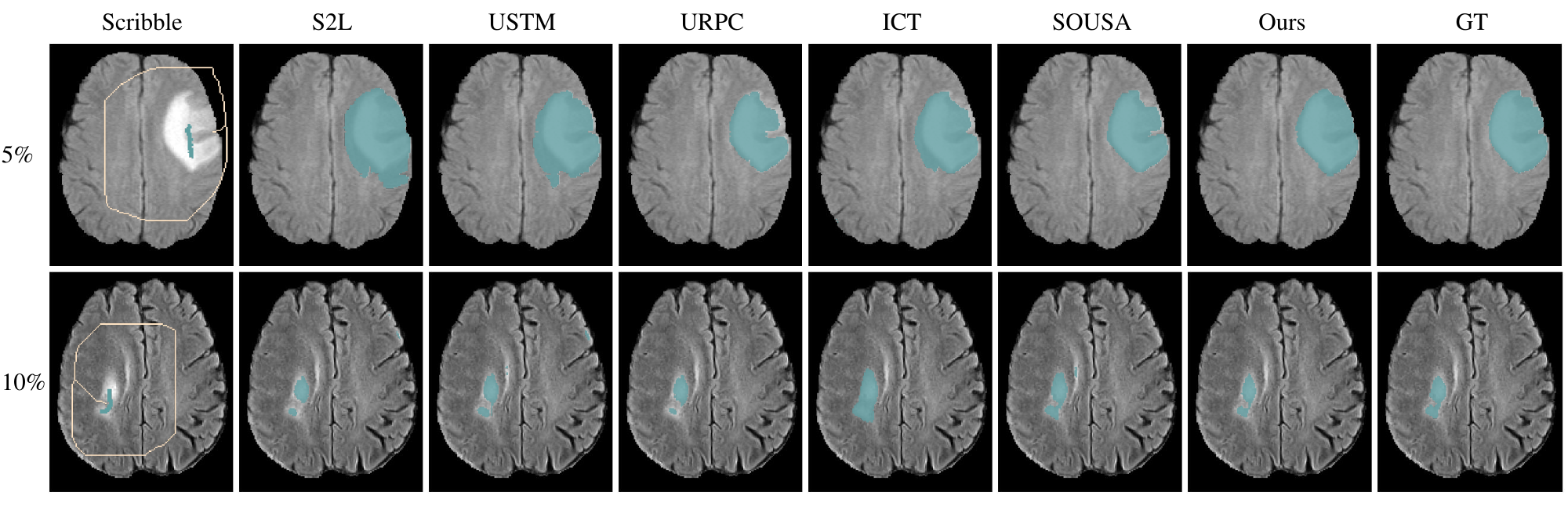} 

\caption{Visual comparison of weakly semi-supervised segmentation results for the BraTS2019 datasets under different labeled ratios.}
\label{fig4}
\end{figure*}

\textbf{BraTS2019 dataset} Segmenting whole brain tumors is a more challenging task due to the inherent ambiguity and diverse shapes of tumors. As shown in Table \ref{table2}, all methods show improvements over the lower bound, but none of them reported better results than the upper bound (0.7848/5\% labeled ratio and 0.8093/10\% labeled ratio). The results indicate that it is challenging to segment the whole tumors with scribble annotations, especially at a 5\% labeled ratio with a gap of 0.0342 Dice score from the upper bound. However, our approach reaches a competitive result (0.8064 Dice and 15.47 95HD) with the upper bound when using a 10\% labeled ratio through superpixel propagation. The visualization of the comparison results is given in Figure \ref{fig4}. Our method can better delineate the tumors than other WSSS methods.

\subsubsection{Results of semi-supervised experiments}
Our SP${ }^3$ learning can be generalized to pure weakly supervised and semi-supervised segmentation. In this study, we use our superpixel propagation idea for semi-supervised segmentation without scribble expansion, replacing sparse annotations with dense annotations and refining pseudo labels of unlabeled data by superpixels. The comparison result with leading semi-supervised methods was listed in Table \ref{table3}. Our method also achieves optimal results compared to the existing semi-supervised method URPC, ICT, TCSM, and UAMT, demonstrating the effectiveness of superpixel propagation in enhancing learning from unlabeled data. 

\begin{table}[!t]
\caption{Comparison results of semi-supervised segmentation on ACDC dataset with 20\% labeled ratio.}
\scalebox{0.9}{
\begin{tabular}{c|cccc}
\toprule
Method             & Dice          & JI            & 95HD       & ASD       \\ \midrule
Dense   Anno. & 0.7579±0.0540 & 0.6511±0.0636 & 12.10±3.63 & 2.53±0.96 \\
UAMT \cite{47}   & 0.8522±0.0271 & 0.7568±0.0381 & 8.17±2.19  & 2.32±0.53 \\
TCSM \cite{43}  & 0.8588±0.0261 & 0.7669±0.0374 & 7.82±1.79  & 1.90±0.53 \\
ICT \cite{ICT}  & 0.8675±0.0246 & 0.7776±0.0352 & 6.68±2.16  & 1.66±0.60 \\
URPC \cite{18}    & 0.8611±0.0325 & 0.7699±0.0453 & 6.70±2.56  & 1.47±0.57 \\
\textbf{Ours}               & \textbf{0.8800±0.0258} & \textbf{0.7965±0.0372} & \textbf{5.56±2.00}  & \textbf{1.32±0.43} \\ \bottomrule
\end{tabular}\label{table3}}
\end{table}
\subsubsection{Results of weakly-supervised experiments}
Our proposed method can also serve as an effective approach for pure weakly supervised learning. As shown in Table \ref{table4}, our method is compared with advanced weakly supervised approaches based on regularization and pseudo-labeling and achieves the best experimental results with 0.8751 Dice. Compared to other pseudo-labeling methods like SS and S2L, our pseudo-labeling method based on superpixel propagation exhibits performance superiority. Furthermore, our method achieves comparable performance to 0.9013 Dice of fully supervised learning using only scribble annotations (FS).
\begin{table}[!t]
\caption{Comparison results of weakly supervised segmentation on ACDC dataset. pCE and FS indicate using all training data with scribble annotations and dense annotations, respectively.}
\centering
\scalebox{0.9}{
\begin{tabular}{c|cccc}
\toprule
Method          & Dice          & JI            & 95HD        & ASD          \\ \midrule
pCE \cite{7}            & 0.6168±0.0363 & 0.4643±0.0365 & 119.10±7.14 & 52.43±6.30 \\
S2L \cite{17} & 0.8386±0.0272 & 0.7336±0.0375 & 25.62±8.53  & 7.21±2.11  \\
USTM \cite{19}  & 0.7695±0.0314 & 0.6411±0.0389 & 80.82±11.08 & 25.05±3.65   \\
RLoss \cite{RLoss}& 0.8743±0.0185 & 0.7844±0.0284 & \textbf{5.75±1.57}   & \textbf{1.45±0.35}    \\
SS \cite{18}  & 0.872±0.077   & -             & 9.9±21.0    & -            \\
\textbf{Ours}        & \textbf{0.8751±0.0165} & \textbf{0.7859±0.0245} & 5.76±2.29   & 1.65±0.60    \\ \midrule
FS              & 0.9013±0.0176 & 0.8276±0.0273 & 4.74±1.77   & 1.25±0.38    \\ \bottomrule
\end{tabular}\label{table4}}

\end{table}

\begin{figure*}[t]
\centering
\includegraphics[width=0.85\paperwidth]{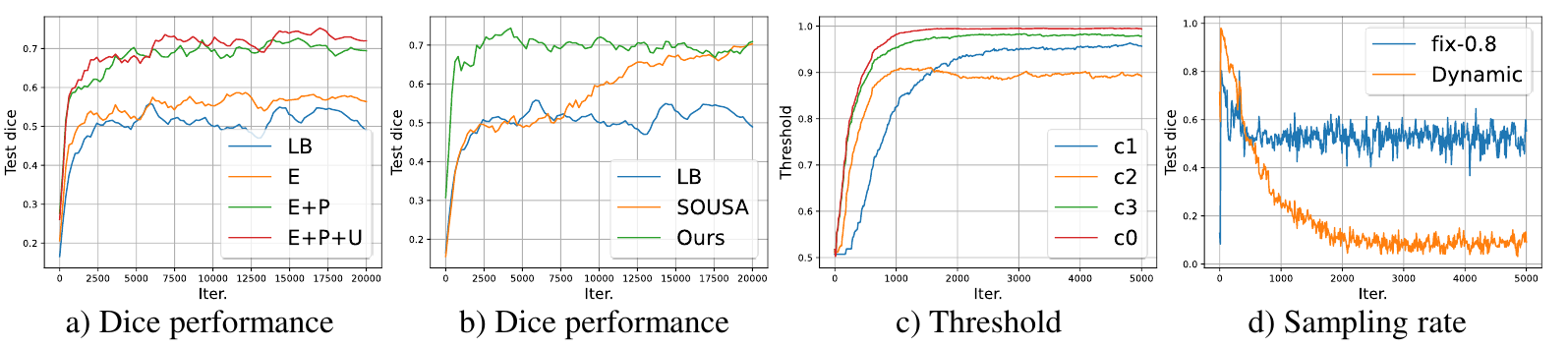} 
\caption{Quantitative analysis in iteration. We present the training curves on fold1 of the ACDC dataset with a 10\% annotation ratio as follows: a) Dice performance of ablation experiments. LB represents the lower bound, E corresponds to E\_scri, P corresponds to Pseu, U corresponds to Unce. b) Dice performance of comparative methods. c) Dynamic threshold of different class. d) Class-average superpixel sampling rate, representing the proportion of superpixels selected based on the dynamic threshold.}\label{fig5}

\end{figure*}

\begin{figure*}[]
\centering
\includegraphics[width=0.80\paperwidth]{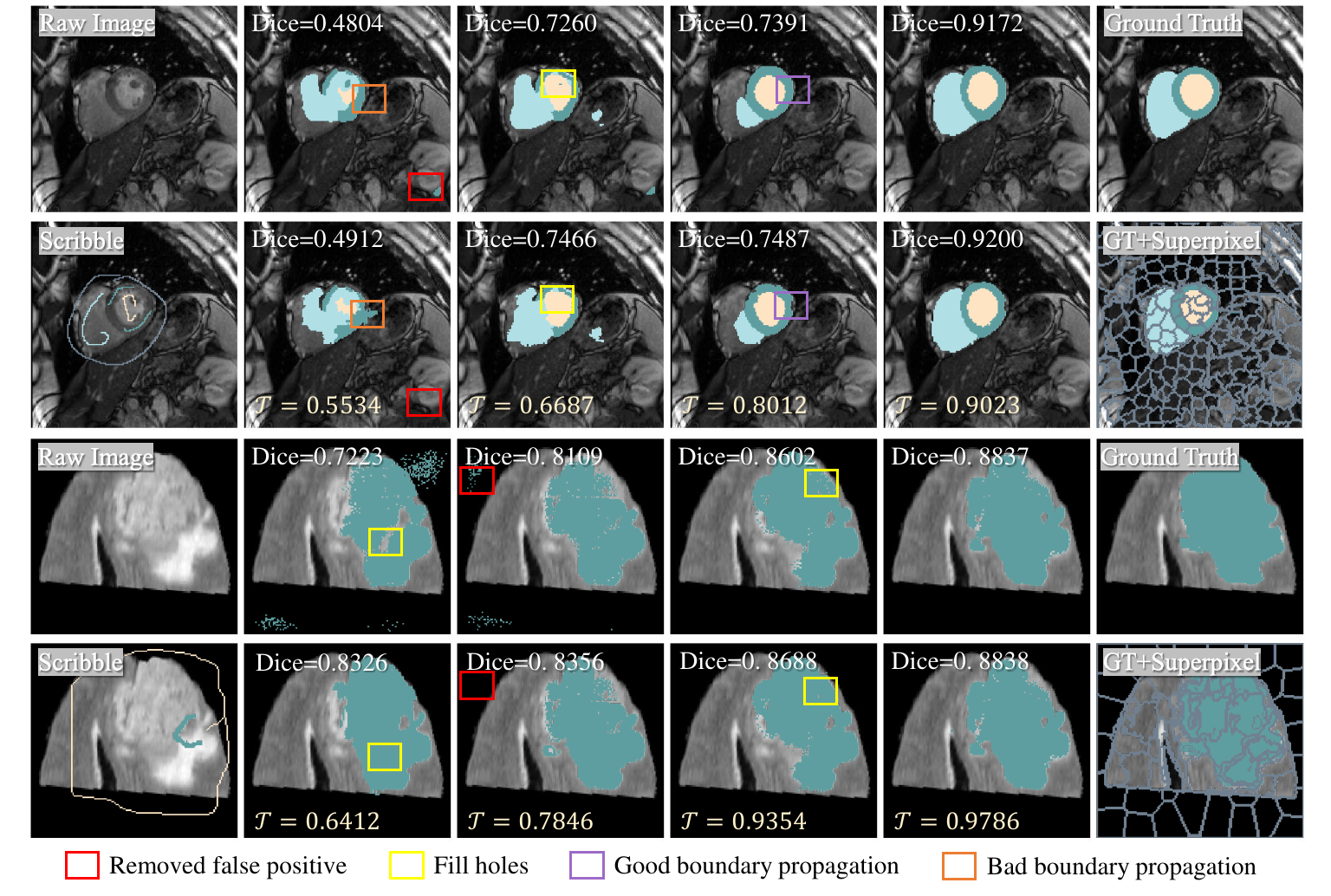} 
\caption{Qualitative analysis of pseudo label refinement during iteration on two datasets. Upper rows is the original predictions and the lower rows show the refined pseudo label. $\mathcal{T}$ is the class-averaged threshold.}
\label{fig6}
\end{figure*}
\subsection{Ablation experiments}
To concretely validate the practicality of each module, we conducted ablation experiments on two datasets in two scenarios: using only scribble data $D_ l$ and incorporating unlabeled data ($D_ l$+$D_ u$) for training. Compared to using only scribble data, we observe that the employment of expanded scribbles for partial supervision led to an improvement over LB, particularly for tumor segmentation (improved by 0.0838 Dice). The pseudo-label supervision refined by superpixels further enhanced segmentation performance, especially the boundary accuracy on the ACDC dataset (reduced by 25.14 ASD). When incorporating unlabeled data using refined pseudo-labels, we can observe significant performance improvement (0.0684 Dice on the ACDC dataset and 0.0664 Dice on the BraTS dataset) brought by the rich sample information from the utilization of unlabeled data. Moreover, with the guidance of superpixel-level uncertainty, the network achieves more efficient and stable information extraction of labeled and unlabeled data and further optimizes the segmentation performance. 
\begin{table}[]
    \caption{Results of ablation experiments on the with dataset under 20\% labeled ratio and the BraTS dataset with 10\% labeled ratio. E\_scri represents the utilization of $\mathcal{L}_{sup}$ with expanded scribbles, Pseu represents the supervision with refined pseudo-label, Unce indicates the guidance of superpixel-level uncertainty, and $D_u$ represents the usage of unlabeled data.}
    \centering
    \begin{tabular}{cccc|cccc}
    \toprule
                              &                        &                        &                        & Dice                       & ASD                      & Dice                       & ASD                       \\ 
    \multirow{-2}*{E\_scri} & \multirow{-2}*{Pseu} & \multirow{-2}*{Unce} & \multirow{-2}*{$D_u$} & \multicolumn{2}{c}{ACDC-20\%} & \multicolumn{2}{c}{BraTS-10\%} \\ \midrule
    \multicolumn{4}{c|}{Sparse Anno.}                                                                                                                                                      & 0.5774                     & 43.93                   & 0.6042                     & 32.26                     \\ \midrule
    \ding{51}                                                 &                                                &                                                &                                                & 0.6362                     & 34.47                    & 0.6880                     & 14.83                     \\
    \ding{51}                                                 & \ding{51}                                              &                                                &                                                & 0.7290                     & 9.33                     & 0.7185                     & 10.54                     \\
    \ding{51}                                                 & \ding{51}                                              & \ding{51}                                              &                                                & 0.7363                     & 7.25                     & 0.7387                     & 10.39                     \\
    \ding{51}                                                 & \ding{51}                                              &                                                & \ding{51}                                              & 0.7974                     & 3.53                     & 0.7849                     & \textbf{4.44}                      \\
    \ding{51}                                                 & \ding{51}                                              & \ding{51}                                              & \ding{51}                                              & \textbf{0.8186}                     &\textbf{ 2.26}                     & \textbf{0.8064}                     & 4.72                     \\ \bottomrule
    \end{tabular}
    \end{table}

\section{Further analysis}
\subsection{Qualitative analysis in iteration}
To analyze why and how our method works, both quantitative and qualitative analyses are presented as follows:

\textbf{Convergence:} In Figure \ref{fig5}(a), we explored the effects of different modules with only $D_ l$ for training. We can observe that supervision with expanded scribbles and refined pseudo-labels both improves the convergence speed. This is primarily due to regional information provided by expanded scribble, and the supervision of the entire image region by the complete pseudo-label facilitating the learning of complete structural appearance. The supervision guided by superpixel-level uncertainty also contributes to more stable early-stage learning and aids in improving performance by focusing on high-quality regions during training. In Figure \ref{fig5}(b), our method converges faster than the consistency-based SOUSA method, since our method accelerates segmentation learning by providing denser labels directly.

\textbf{Dynamic thresholding:} We present the comparison of threshold and sampling rates in iteration in Figures \ref{fig5}(c) and \ref{fig5}(d). At the beginning of training, the threshold is set low to introduce as much potentially accurate superpixel information as possible. As performance improves, the threshold adaptively increases to filter out noisy superpixels and reduce biases. Additionally, we illustrate the sampling rate with a fixed threshold as a reference. The fixed threshold corresponds to a sampling rate that quickly converges to around 50\%. The dynamic threshold adjusts based on the overall learning state with the sampling rate converging to 10\% ultimately.

\subsection{Quantitative analysis}
\textbf{Dynamic thresholding:} To further demonstrate the importance of dynamic thresholding, we conduct a comparative study on thresholding strategies, as shown in Table \ref{tab6}. Compared with using pseudo label without superpixel-based refinement (pseu w/o SP), directly assigning maximum class (w/o $\mathcal{T}$) degrades the performance, as it introduces superpixel noise and it’s hard to correct biases during learning. Using a fixed threshold 0.8 ($\mathcal{T}$=0.8) to filter potentially high-quality superpixels for refinement can attain performance improvement over directly using pseudo label without refinement, and adjusting threshold with EMA ($\mathcal{T}_t$) can get higher performance. These results prove that screening superpixels can reduce the influence of bad superpixels and improve the segmentation performance. Our class-specific dynamic threshold ($\mathcal{T}_t (c)$) achieves the best performance, facilitating better propagation of beneficial superpixel information with the consideration of category difference.

\begin{table}[h]
\centering
\caption{Comparison of the thresholding strategy on ACDC dataset under 10\% labeled ratio.}
\begin{tabular}{ccc }
\toprule
Threshold   & Dice   & ASD  \\ \midrule
pseu w/o SP & 0.7503 & 3.89 \\
w/o $\mathcal{T}$       & 0.7290 & 7.63 \\
$\mathcal{T}$=0.8       & 0.7625 & 6.13 \\
$\mathcal{T}_t$        & 0.8066 & 4.52 \\
$\mathcal{T}_t(c)$     & 0.8186 & 2.85 \\ \bottomrule
\end{tabular}\label{tab6}
\end{table}

\textbf{Refinement effect:} The refinement with superpixels filtered by the dynamic threshold can help the model generate better pseudo labels during iterations, as depicted in Figure \ref{fig6}. As the iteration proceeds, the generated pseudo-labels get closer to the ground truth. Utilizing superpixels helps to eliminate false positive predictions and fills holes in pseudo labels. By employing dynamic thresholds for superpixel selection, we filter out bad superpixels and only use higher-quality superpixels for refinement, facilitating the propagation of favorable boundary information throughout the network. When the accuracy of pseudo-labels increases, the dynamic threshold also increases continuously causing fewer relabeling operations.

\textbf{Superpixel-level uncertainty}
To deeply elucidate the motivation behind our uncertainty strategy, we also evaluate different strategies to generate uncertainty maps and show the results in Table \ref{tab7}. Firstly, we investigate the level of uncertainty map. Compared with the pixel-level uncertainty map \cite{36} generated by our same prediction consistency (Pixel-level), the superpixel-level uncertainty map helps gain better results. This improvement is attributed to the fact that the superpixel-level evaluation aggregating information from pixels within superpixels ensures consistency between the evaluation level and refinement level, leading to a more reliable assessment of refined pseudo-labels. Secondly, we explore the measurement of uncertainty. We calculate the uncertainty by averaging the commonly used KL divergence \cite{73} between two soft outputs within the superpixel to obtain a KL-based superpixel-level uncertainty map (KL\_SP). Our method achieves more comprehensive results by utilizing the proportion of inconsistent predictions of each superpixel as superpixel-level uncertainty. Additionally, we reverse the magnitude relation between the uncertainty and the weight (Ours\_re), assigning higher weights to the regions with higher uncertainty (with the weight denoted as $w_{i j}=e^{-u_j+1}$), which significantly declines the performance. This decline is primarily due to the impact of substantial noise related to high-uncertainty regions. Our superpixel-level uncertainty guidance helps the network focus on more reliable regions and achieve the best result, leading to robust network learning.

\begin{table}[h]
\centering
\caption{Comparison of the uncertainty strategy on ACDC dataset under 10\% labeled ratio.}
\begin{tabular}{ccc}
\toprule
Uncertainty & Dice   & ASD   \\ \midrule
w/o $U$       & 0.7974 & 3.53  \\
Pixel-level   & 0.8010 & 2.94  \\
KL\_SP      & 0.7946 & 2.33  \\
Ours\_re    & 0.7326 & 16.26 \\
Ours        & 0.8186 & 2.85  \\ \bottomrule
\end{tabular}\label{tab7}
\end{table}

\textbf{Hyperparameters of superpixel generation}
In this study, we employ the conventional superpixel generation algorithm, SLIC \cite{50}, for image data mining, a technique often used in domain adaptation-related research \cite{51,52,53}. In experiments, we utilize the scikit-image package to perform 2D and 3D SLIC segmentation on the ACDC and BraTS2019 datasets. In our superpixel propagation method, the size of superpixels should be smaller than the target size. Since the size of superpixels is determined by the number of superpixels $n$ and impacts the effect of re-labeling, we use hyperparameter experiments to choose the appropriate value of $n$.

\begin{figure}[t]
\centering
\includegraphics[width=.9\columnwidth]{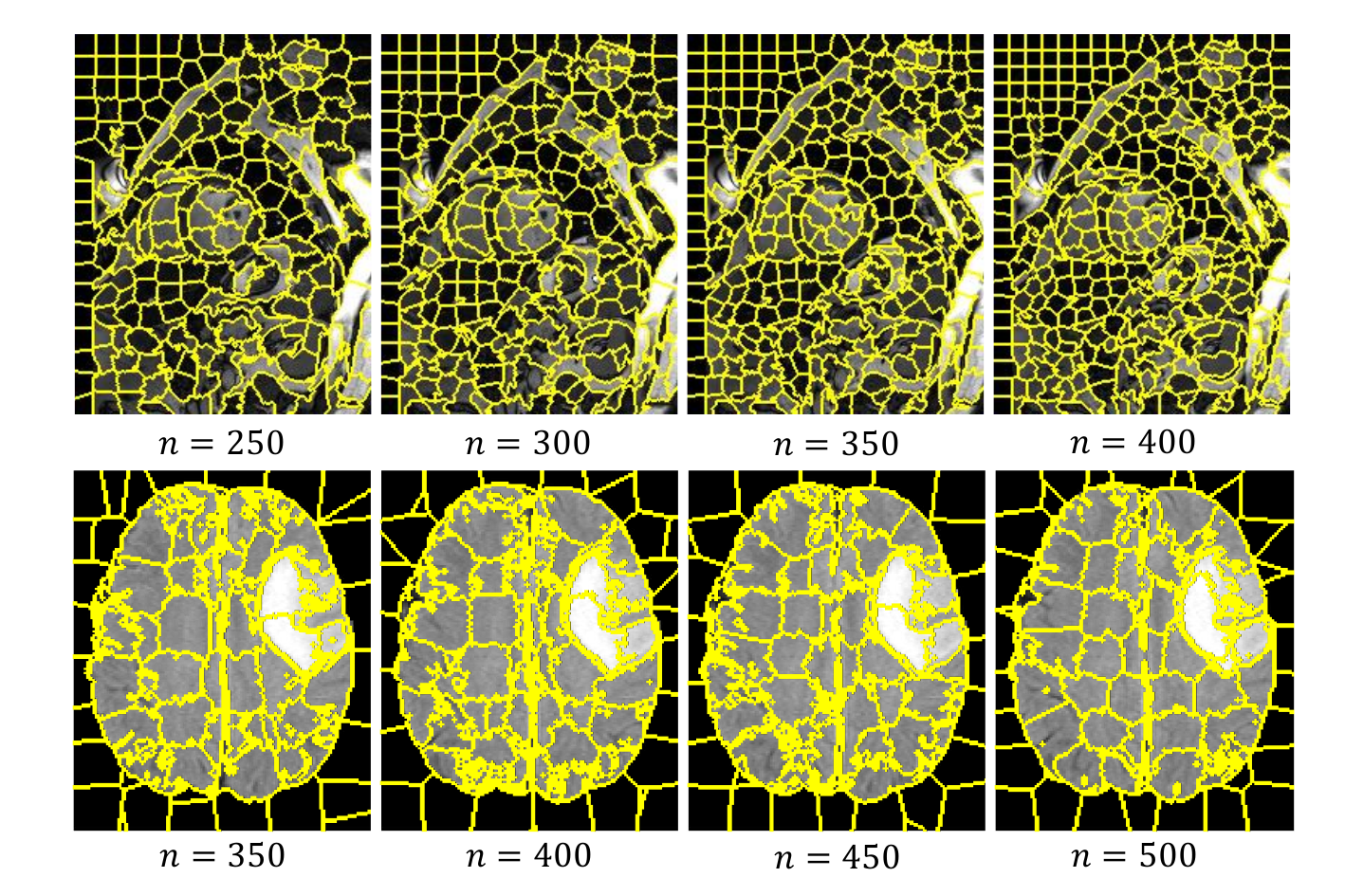} 
\caption{Superpixel segmentation results with different parameters $n$. $n$ represents the number of superpixels in an image or volume.}
\label{fig7}
\end{figure}

We respectively set four sets of hyperparameters for the segmentation on the 5th-fold of ACDC and the BraTS2019 dataset, recording the results in Table \ref{tab8}. The corresponding visualization of superpixel segmentation results is shown in Figure \ref{fig7}. Smaller $n$ results in larger superpixels while larger $n$ produces smaller superpixels. First, we use a fixed threshold of 0.8 to examine the influence of superpixel size on segmentation. The ACDC dataset and BraTS2019 dataset achieve optimal experimental results at the number of superpixels 350 and 400, respectively, so we chose them as the corresponding hyperparameters. In the hyperparametric experiment, when $n$ is small, larger superpixels may lead to false expansion outside the target region when propagating scribble-label. Conversely, when $n$ is large, smaller superpixels with more boundaries cause a decrease in the proportion of semantically valuable boundaries related to the target segmentation, which hinders the propagation of boundary information. To further explore the impact of dynamic thresholds, we conduct experiments with the dynamic threshold strategy (with $\tau_0=0.5$)  to examine its effect on hyperparameters. When using dynamic thresholds, the performance of different hyper-parameters both improved, demonstrating the effectiveness of dynamic thresholding, which makes the network more robust to the hyperparameters.

\begin{table}[]
\centering
\caption{Dice performance of superpixel hyperparameter experiments on 5-th fold of ACDC-20\% and on BraTS-10\%. The red color represents the dice improvement using dynamic thresholding compared to fixed thresholding.}
\begin{tabular}{cccccc}
\toprule
Dataset                 & Threshold & 250                           & 300                           & 350                           & 400                           \\ \midrule
                        & Fix-0.8   & 0.7834                        & 0.7916                        & 0.8078                        & 0.8047                        \\
                        & Dynamic   & 0.7861                        & 0.8024                        & 0.8112                        & 0.8064                        \\
\multirow{-3}{*}{ACDC}  & $\triangle$         & 0.0027 & 0.0098 & 0.0034 & 0.0016 \\ \midrule
Dataset                 & Threshold & 350                           & 400                           & 450                           & 500                           \\ \midrule
                        & Fix-0.8   & 0.7825                        & 0.8036                        & 0.7969                        & 0.7802                        \\
                        & Dynamic   & 0.8026                        & 0.8064                        & 0.8036                        & 0.7989                        \\
\multirow{-3}{*}{BraTS} & $\triangle$         & 0.0201 & 0.0038 & 0.0067 & 0.0187 \\ \bottomrule
\end{tabular}\label{tab8}
\end{table}

\textbf{Annotation costs}
In the experiments, we calculate the annotation costs corresponding to different annotation-efficient learning compared with fully supervised learning, as shown in Table \ref{tab9}. It can be observed that when using 10\% of the labeled data, semi-supervised learning is capable of reducing the annotation cost to around 1/10, while weak annotation reduces the number of annotated pixels to approximately one-third. The weakly semi-supervised method we employ manages to reduce the annotation cost to roughly 3\% of fully supervised learning, demonstrating significant annotation efficiency.

\begin{table}[]
\centering
\caption{Annotation cost of different learning. We use the number of labeled pixels in fully supervised learning as the benchmark of 100\% annotation. We then calculate the ratio of scribble-annotated pixels of all training data in weakly supervised learning, densely annotated pixels of 10\% training data in the semi-supervised scenario, and scribble-annotated pixels of 10\% training data in the weakly semi-supervised scenario, compared to the fully supervised learning.}
\begin{tabular}{ccccc}
\toprule
Dataset & Semi    & Weakly  & Weakly semi & FS    \\ \midrule
ACDC    & 12.91\% & 27.89\% & 3.57\%      & 100\% \\
BraTS   & 9.03\%  & 37.16\% & 3.49\%      & 100\% \\ \bottomrule
\end{tabular}\label{tab9}
\end{table}

\section{Conclusion}
This work mainly focuses on automated medical image segmentation under annotation scarcity. Weakly semi-supervised segmentation (WSSS) is an effective way to alleviate this dilemma, which trains a model with only a small number of scribbles and a large amount of unlabeled data. The proposed superpixel-propagated pseudo-label (SP$^3$) learning is a novel WSSS method that achieves promising results with high label efficiency. Specifically, the structural information contained in superpixels complements insufficient supervision from limited scribbles by expanding the scribbles to provide denser labels. The pseudo-label learning handles the inconsistent supervision caused by mixed supervision, and the superpixel refinement and the superpixel-level uncertainty are utilized to ensure stable pseudo-label supervision. Experimental results show that our method outperforms the current state-of-the-art methods significantly on two public datasets in the WSSS setting, demonstrating its remarkable annotation efficiency. Besides, our method shows promising results in pure weak and semi-supervised settings. These results indicate its substantial potential to serve as a general solution for other annotation-efficient medical segmentation tasks and may inspire research on other learning paradigms, such as partially supervised learning and weakly supervised learning with points \cite{li2023multi}.

\section*{Acknowledgments}
This research was supported by National Natural Science Foundation of China (Grant Nos. 82072021) and Shanghai science and technology innovationaction plan (Grant Nos. 19511121302).


%

\bibliographystyle{IEEEtran}
\bibliography{ieee_ref}
\newpage

\vfill

\end{document}